\documentclass[letterpaper, 10 pt, conference]{ieeeconf}
\IEEEoverridecommandlockouts
\usepackage{cite}
\usepackage{amsmath,amssymb,amsfonts}
\usepackage{algorithmic}
\usepackage{graphicx}
\usepackage{textcomp}
\usepackage{xcolor}
\usepackage{subcaption}
\usepackage{mathtools}
\usepackage{scalerel}
\usepackage{siunitx}
\usepackage{booktabs}
\usepackage{float}
\usepackage{hyperref}
\usepackage[keeplastbox]{flushend}
\def\BibTeX{{\rm B\kern-.05em{\sc i\kern-.025em b}\kern-.08em
    T\kern-.1667em\lower.7ex\hbox{E}\kern-.125emX}}

\title{\LARGE \bf
Locating Transparent Objects to Millimetre Accuracy}

\author{Nicholas Adrian and Quang-Cuong Pham% <-this % stops a space
  \thanks{The authors are with the School of Mechanical and Aerospace
    Engineering, Nanyang Technological University, Singapore.}%
}

\begin{document}

\maketitle
\thispagestyle{empty}
\pagestyle{empty}

\begin{abstract}
  Transparent surfaces, such as glass, transmit most of the visible
  light that falls on them, making accurate pose-estimation
  challenging. We propose a method to locate glass objects to
  millimetre accuracy using a simple Laser Range Finder (LRF) attached
  to the robot end-effector. The method, derived from a physical
  understanding of laser-glass interactions, consists of (i) sampling
  points on the glass border by looking at the glass surface from an
  angle of approximately 45 degrees, and (ii) performing Iterative
  Closest Point registration on the sampled points. We verify
  experimentally that the proposed method can locate a transparent,
  non-planar, side car glass to millimetre accuracy.
\end{abstract}

\section{Introduction}
\label{part:introduction}

Transparent surfaces, such as glass, transmit most of the visible
light that falls on them. For a common plate of glass, only about 4\%
of the incident light are \emph{reflected} from each surface while the
other 92\% are \emph{transmitted} across. This non-opaque
characteristic prevents common vision sensors from locating
transparent objects with high accuracy. Given the prevalence of
transparent objects in industrial and home environments, such a
limitation hinders the application of many vision-based robotics
systems.

Most robotics research into the detection/recognition of transparent
objects is only concerned with coarse localisation -- typically,
centimetre-accuracy in indoor navigation or
tens-of-centimetre-accuracy in outdoor navigation. By contrast,
industrial tasks (transport, assembly, glue dispensing on car glasses,
etc.) require a much higher level of accuracy, typically of several
millimetres or submillimetre. Furthermore, most of the existing
methods in transparent objects shape and pose estimation require
elaborate apparatus setup (e.g. specific and fixed configurations of
multiple sensors).

Here, we propose a method to locate glass objects to millimetre
accuracy using a simple LRF attached to the robot
end-effector. The proposed method can locate both planar and
non-planar objects for subsequent manipulation. Briefly, the proposed
localisation and manipulation pipeline is as follows:

\begin{enumerate}
\item \emph{Scanning}: Sample $n$ points along the object border by
  looking at the surface with the LRF from an angle of approximately
  45 degrees. We assume in this paper that a coarse localisation has
  been performed beforehand, using existing
  methods\cite{Hata1996}\cite{Ben-Ezra2003};
\item \emph{Registration}: Perform Iterative Closest Point (ICP)
  registration of the $n$ points against an available 3D model;
\item \emph{Manipulation}: Based on the object pose estimate, compute
  the robot trajectory to perform the manipulation (here: following
  the object contour mimicking a glue dispensing task).
\end{enumerate}

The remainder of the paper is organized as follows. In Section
\ref{part:previouswork}, we discuss several works done in localisation
of transparent objects. In Section \ref{part:laserandglass}, we survey
the background knowledge on laser-glass interactions. In Section
\ref{part:pipeline}, we present our glass localisation
method. Sections~\ref{part:experimentA} and~\ref{part:experimentB}
report experimental results on respectively point and full-object
localisation. Finally, in Section \ref{part:conclusion}, we conclude
and sketch directions for future work.

\begin{figure}[t]
  \centering
  \includegraphics[width=\columnwidth]{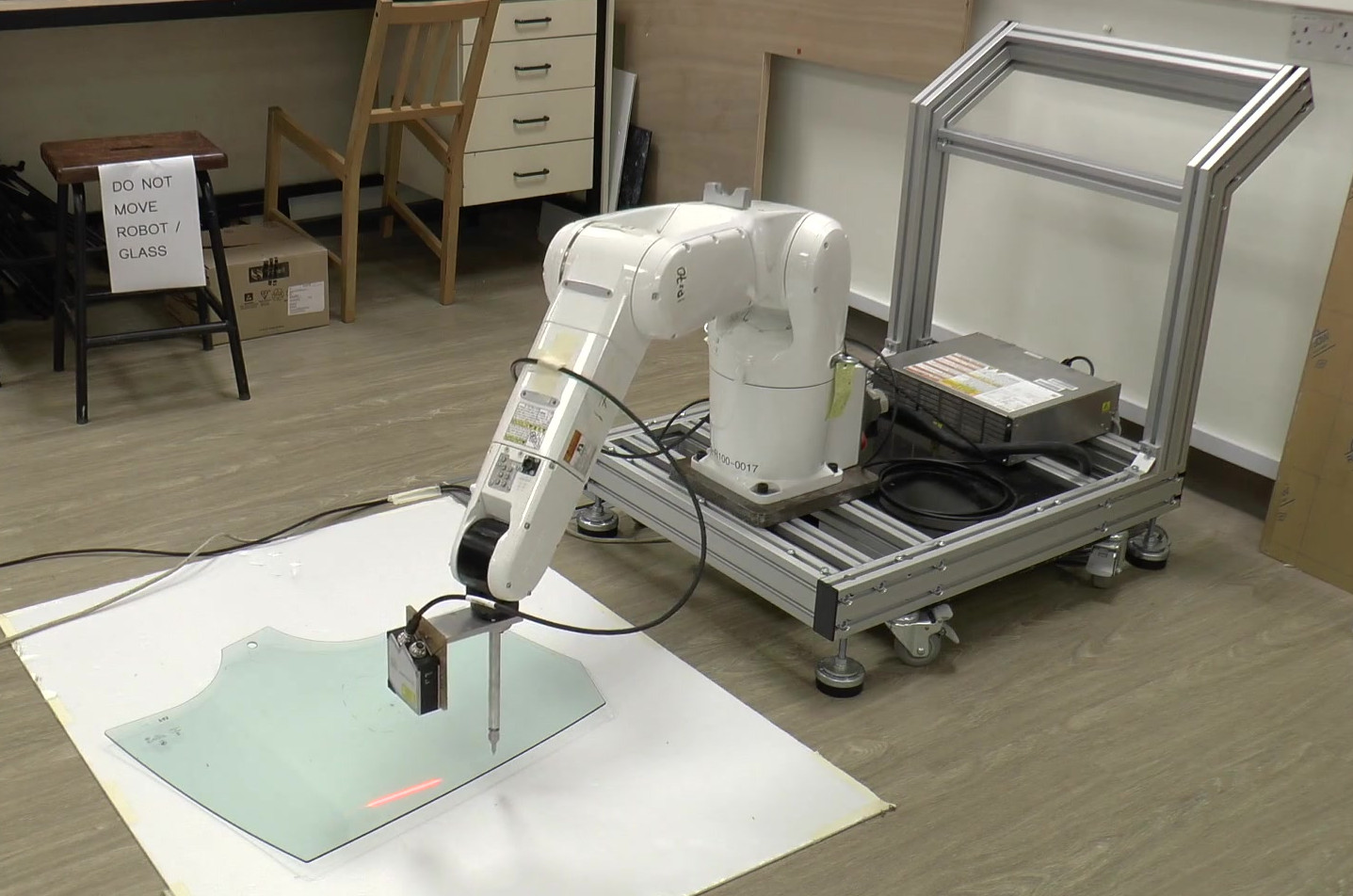}
  \caption{Experimental setup, comprising a 6-DOF robotic arm, a 2D
    LRF, and a nozzle-like tooltip attached to the robot flange. The
    object to be located is an actual non-planar side car glass.
  Video demonstration of the full pipeline is available at \href{https://youtu.be/QoEyQtwsckE}{https://youtu.be/QoEyQtwsckE}.}
  \label{fig:fig1}
\end{figure}

\section{Previous work on localisation of transparent objects}
\label{part:previouswork}

Most research into transparent object recognition does not require the
estimated object to be manipulated upon. As such, the estimated models
are often simplified and lack the necessary detail for accurate
manipulation. Furthermore, most of the methods require elaborate
apparatus setup.

In \cite{Hata1996}, Hata used a structured light setup to project
stripe patterns into transparent object with one flat side. The
refracted pattern is then inspected and extracted using imaging
sensor. Genetic Algorithm is then iteratively employed to construct a
3D model until the calculated error is within a threshold.
The author achieved shape accuracy below 20\% after running 50
generations for six hours.

In \cite{Ben-Ezra2003}, the authors tracked features from different frames
of a moving camera to estimate shapes and poses of transparent objects.
The shape reconstruction assumes parametric form of the transparent
object. They showed that the problem can be highly non-linear even for
object with simple shape. Achieved shape parameters error range from
submillimetre to tens-of-centimetre while there was no data on real experimental
pose parameters.

In \cite{Narita2005}, Narita proposed a non-contact optical method for
measuring the 3D shape of transparent objects and their refractive indices using LRF.
The method relies on observing and triangulating the two strongest
reflected light rays. The author could measure thickness of two
transparent object of simple shapes to submillimetre error, obtained
through non-trivial arrangement of multiple sensors.

In \cite{Kutulakos2008}, Kutulakos calculated depth maps of opaque
and transparent specular objects by reconstructing light paths from
a light source. The number of viewpoints required depends on the
number of times the light get redirected. An experiment to reconstruct
a diamond-shaped glass object with eight faces and from five viewpoints
was done. The average RMS distance from the obtained points to the
estimated plane is 1.33 mm.

In the robotics field, transparent object detection is of special
interest for mobile robot localisation especially to avoid
collision. Developments in non-flat object recognition through robotic
arm haptic-based approach are of interest as well, which can be generalized
to transparent objects. These approaches hold promise with their
non-reliance on external vision sensors, which tend to fail transparent
objects.

In \cite{Foster2013}, Foster tackled the problem of mobile robot
localisation in an environment with many glass obstacles using a LIDAR
sensor. The author looked into the behaviour and working of LIDAR
sensors to recognise the visibility of glass at certain view angles
and conditions. These view angles are later experimentally identified
and used in modifying the standard occupancy grid algorithm. The algorithm
looks for evidence that a transparent object occupies a cell but
does not estimate the pose of occupying object.

In \cite{Allen1990}, Allen employed Utah-MIT hand fitted with tactile
sensor pads that was attached to PUMA 560 manipulator. Borrowing
previous research on human haptic system \cite{Klatzky1988}, they
extended human hand movement strategies in discovering 3D objects
attributes to robotic domain. The idea was to generate a rough initial
shape estimate using Grasping by Containment before performing
additional exploratory procedures (EPs) to refine the structure.

In \cite{Ibrayev2004}, Ibrayev developed an invariant-based method to
recover surface geometry with few data collected from touch
sensors. Firstly, curve class recognition is done by calculating
differential invariants for few classes of quadratic curves and
special cubic curves. From there, actual curve is estimated based on
the shape parameterisation as well as contact locations on the curve.
The author claimed an average relative error of 1\% depending on the
curvature and its derivative estimation performance.

\section{Laser-glass interactions}
\label{part:laserandglass}

\subsection{Triangulation-based laser scanners}

\begin{figure}[htp]
  \centering
  \includegraphics[width=\columnwidth]{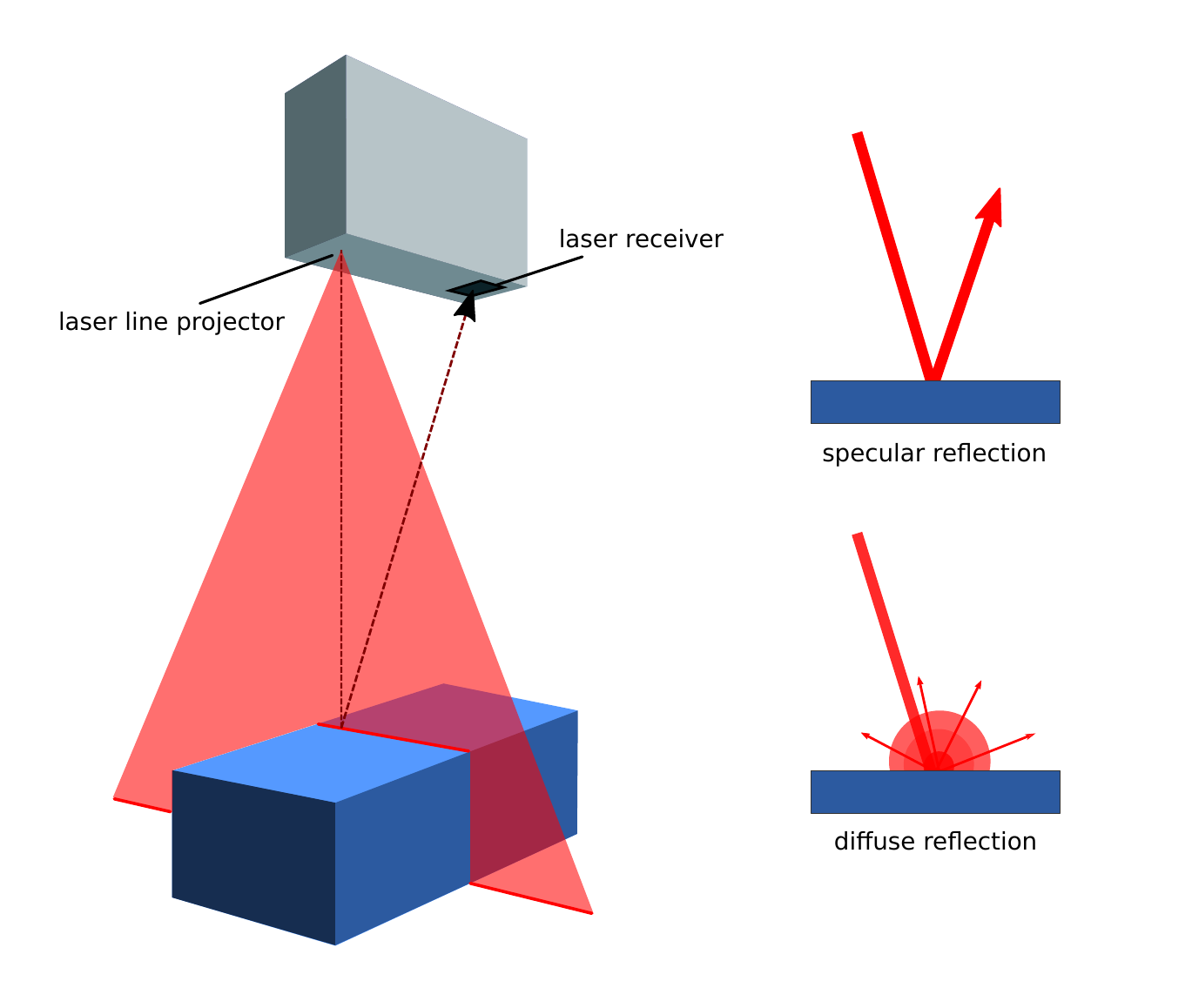}
  \caption{Left: Scan geometry of the triangulation-based laser
    scanner used in this research. Right: Two types of light
    reflection: specular and diffuse.}
  \label{fig:fig2}
\end{figure}

The triangulation-based laser scanner works
by projecting a laser line beam onto the object surface to be
determined. The reflected light is then subsequently captured by the
scanner's receiver. In our application, the scanner projects laser
line onto the object at right angle while the receiver is pointed
towards the reflected line from an angle as seen in Figure
\ref{fig:fig2} (left). In the sensor's receiver, the reflected light will
fall on an array of Charge-Coupled Device (CCD), from which the object
position is calculated.

Most of the light that falls onto the transparent material either gets
reflected or transmitted. The reflected light consists of two different types:
\emph{specular} and \emph{diffuse}, as shown in Figure \ref{fig:fig2}
(right). In the former, the incident and reflected rays' angles are equal but
located on opposite sides of the reflected surface normal. In the
latter, the incident ray is scattered to reflected rays that cover
many angles. Meanwhile the transmitted ray might be transmitted again through
the rear surface or get refracted multiple times internally before
being transmitted out from either surfaces.

\subsection{Laser reflection from glass surfaces}

The laser scanner's ability to register points from glass surfaces
depends on several main factors, two of which are: laser reflection type and
presence of opaque object in the background.

\begin{figure}[htp]
  \centering
  \smallskip\smallskip
  \begin{subfigure}{0.5\columnwidth}
    \centering
    \includegraphics[width=\linewidth]{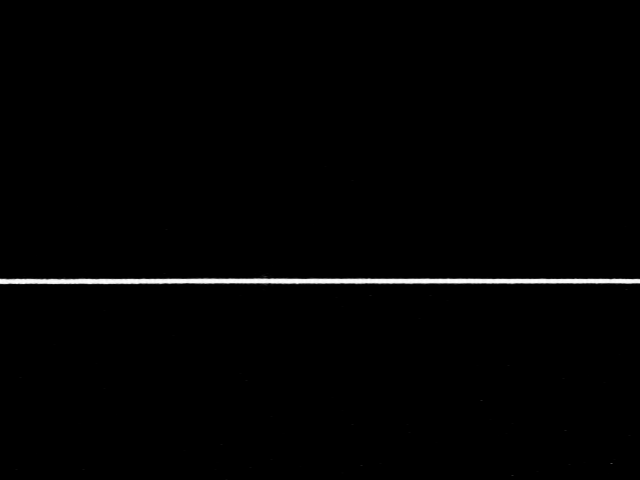}
    \caption{}
    \label{fig:fig3-0a}
  \end{subfigure}%
  \begin{subfigure}{0.5\columnwidth}
    \centering
    \includegraphics[width=\linewidth]{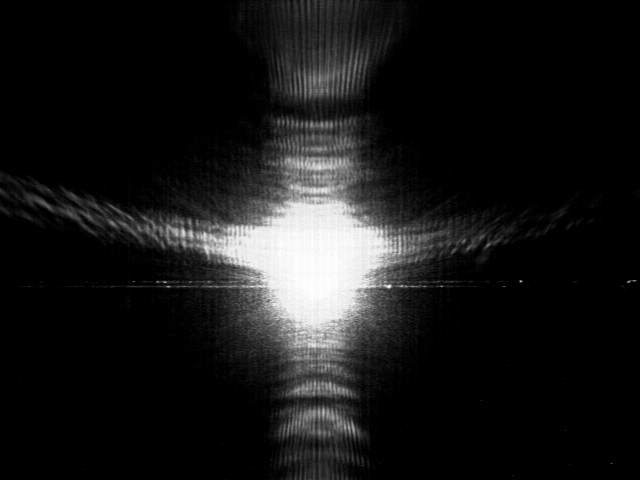}
    \caption{}
    \label{fig:fig3-0b}
  \end{subfigure}
  \caption{Two images from the receiver camera compare (a) normal
    sensor reading from diffuse reflection against (b) saturated
    sensor reading from specular reflection of a same flat contour.}
  \label{fig:fig3-0}
\end{figure}

Specular reflections that reach the receiver sensor tend to saturate
the CCD, causing blooming effect \cite{Amir2017} as shown in Figure
\ref{fig:fig3-0}. The saturated pixels will lead to inaccurate reading
and hence false points being registered. Avoiding specular reflections
can be challenging especially when dealing with non-planar glass with
varying surface normal angles. 

On the other hand, the scanner's receiver relies on diffuse reflection
of the laser line to capture the surface contour. However, the diffuse
reflections from the glass surface tend to be very low. As previously
mentioned, only about 8\% of the laser light get reflected back from
both surfaces. Only part of this is reflected diffusely due to glass
surface imperfection or small particles. Depending on the laser
scanner, some setting adjustment on exposure time or minimum intensity
threshold might be required to pick up these low intensity points.

Presence of opaque surfaces in the background can also cause the
scanner to register false reflections. This problem is sometimes referred
to as the Mixed Pixels problem in literature \cite{Ye2008}. In some cases
involving LIDARs, the range measurement with higher intensity has
a higher chance of being selected \cite{Foster2013}. The author argued that the intensity
of return from glass can overpower the opaque background object only
for certain critical angle near the glass normal for a constant distance
between the two objects. An example given was that when a diffuse wall
was placed 1 m behind the glass, the glass will only be visible from a
small 0.25 degree angle range.

\begin{figure}[htp]
  \centering
  \smallskip\smallskip
  \begin{subfigure}{0.5\columnwidth}
    \centering
    \includegraphics[width=\linewidth]{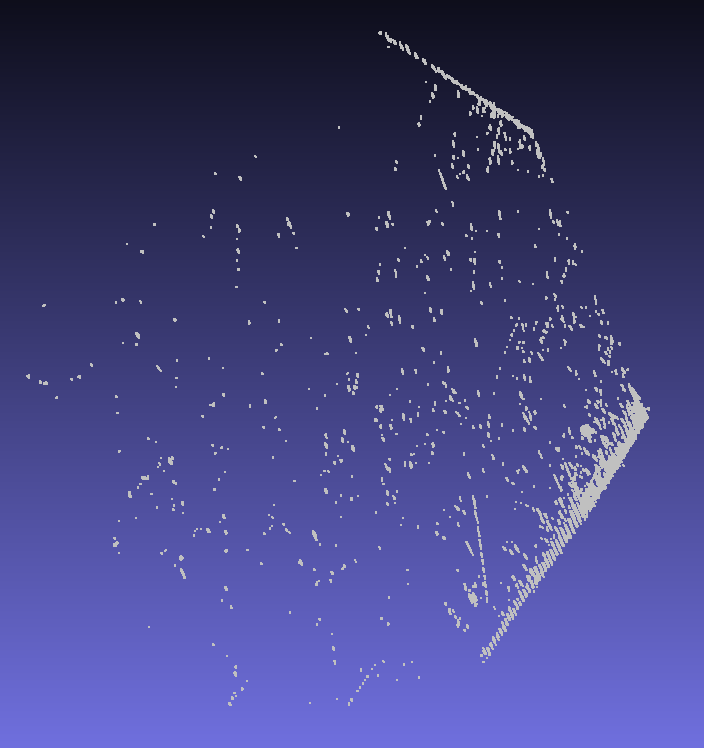}
    \caption{}
    \label{fig:fig3-1a}
  \end{subfigure}%
  \begin{subfigure}{0.5\columnwidth}
    \centering
    \includegraphics[width=0.8\linewidth]{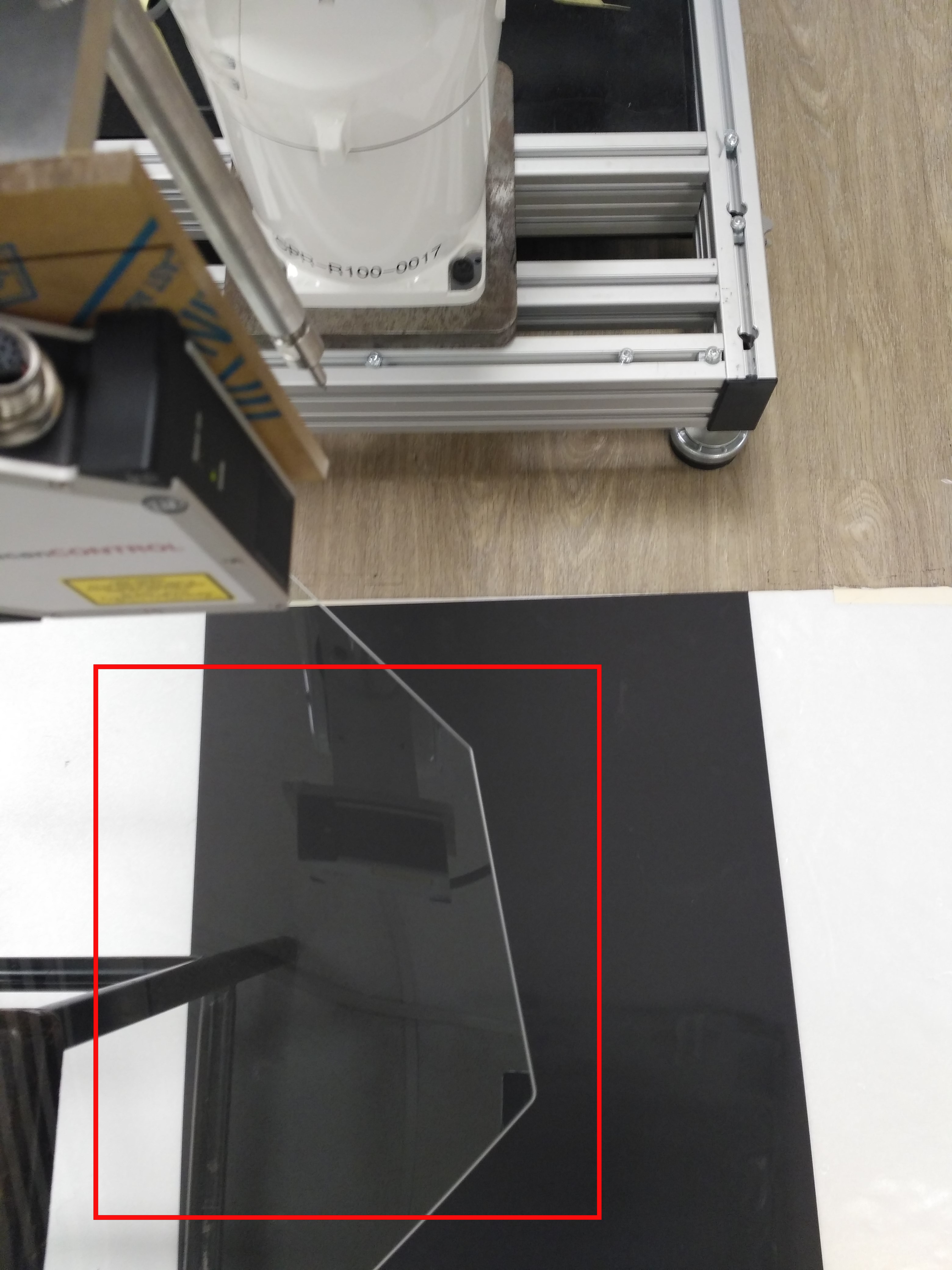}
    \caption{}
    \label{fig:fig3-1b}
  \end{subfigure}
  \caption{(a) Sparse 3D points obtained by scanning along the glass
    surface. Most of the points are reflected and captured due to
    glass imperfection (b) The rectangular scan section of the glass.}
  \label{fig:fig3-1}
\end{figure}

In Figure \ref{fig:fig3-1}, we show the obtained reading from the
glass surface taking the above factors into account. Figure
\ref{fig:fig3-1a} shows the sparse 3D points constructed by moving the
robotic arm to scan along a flat glass surface marked in Figure
\ref{fig:fig3-1b}. The LRF scanning properties used are listed in
Table \ref{tab:1}.

\begin{table}
  \caption{LRF properties}
  \setlength{\tabcolsep}{10pt}
  \centering
  \begin{tabular}{lc}\toprule
    Property & Value \\\midrule
    Angle (from normal) & 2 \\
    Exposure time & 5 ms \\
    Selection method & highest intensity \\
    Distance from glass to opaque surface & 57 cm \\
    \bottomrule
  \end{tabular}
  \label{tab:1}
\end{table}

\begin{figure}[htp]
  \centering
  \smallskip\smallskip
  \begin{subfigure}{0.5\textwidth}
    \includegraphics[width=1\linewidth]{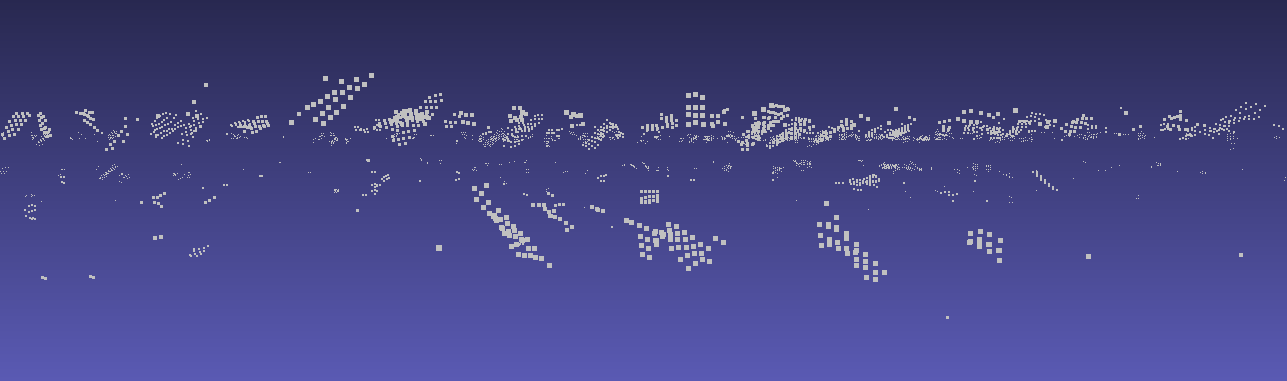}
    \caption{}
    \label{fig:fig3-2a}
  \end{subfigure}
  \begin{subfigure}{0.5\textwidth}
    \includegraphics[width=1\linewidth]{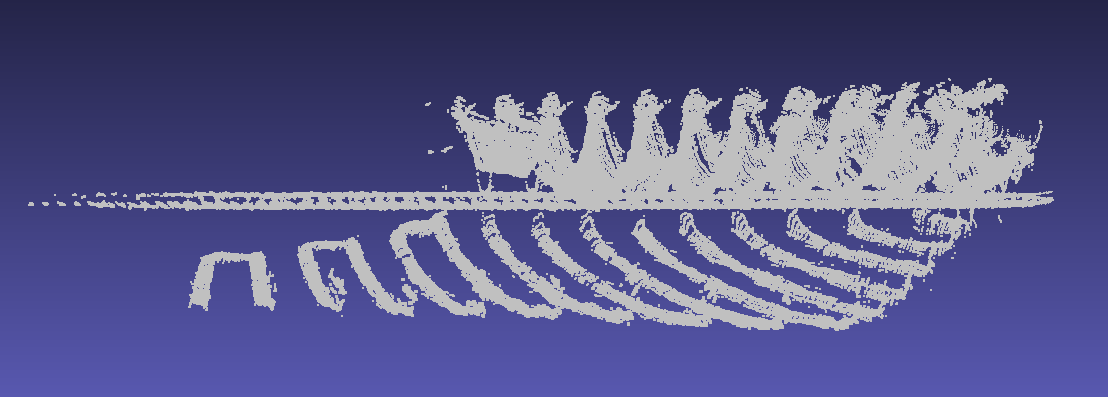}
    \caption{}
    \label{fig:fig3-2b}
  \end{subfigure}
  \caption{(a) Points captured along the top and bottom surfaces of a
    flat glass are highly irregular. (b) Inaccurate reading of a flat
    glass due to specular reflection.}
  \label{fig:fig3-2}
\end{figure}

However, a closer look into the obtained 3D points in Figure
\ref{fig:fig3-1} reveal irregularities along both surfaces of the
glass as shown in Figure \ref{fig:fig3-2a}. The point distribution is
characterised by distinct inclined spikes which mask the true surface
flatness. Outlier points are also registered along the thickness of
the glass at different parts of the glass. In Figure
\ref{fig:fig3-2b}, we show the noisy reading obtained when specular
reflection is involved. The glass panel spans the horizontal middle
section of the image, with irregular points distributed above and
below it. All of these observations point towards the unpracticality
of the scanned output.

In section \ref{part:pipeline}, we propose an alternate method to
realize our glue-dispensing robotic system that does not rely on such
irregular output.

\subsection{Laser reflection from glass edges}

While the transparent surface of the car glass window gives mostly
unreliable readings, the edge offers a better compromise. Most glass
panels, such as the car window, have higher surface irregularities
along the border that can be attributed to applied glass edge
finishing which is often required for aesthetics and safety
reasons. This physical property allows lasers that fall on the less
transparent glass edge to give out more diffuse reflection.
This is desirable as it provides cleaner and hence more
accurate range measurement.

\begin{figure}[htp]
  \centering
  \includegraphics[width=\columnwidth]{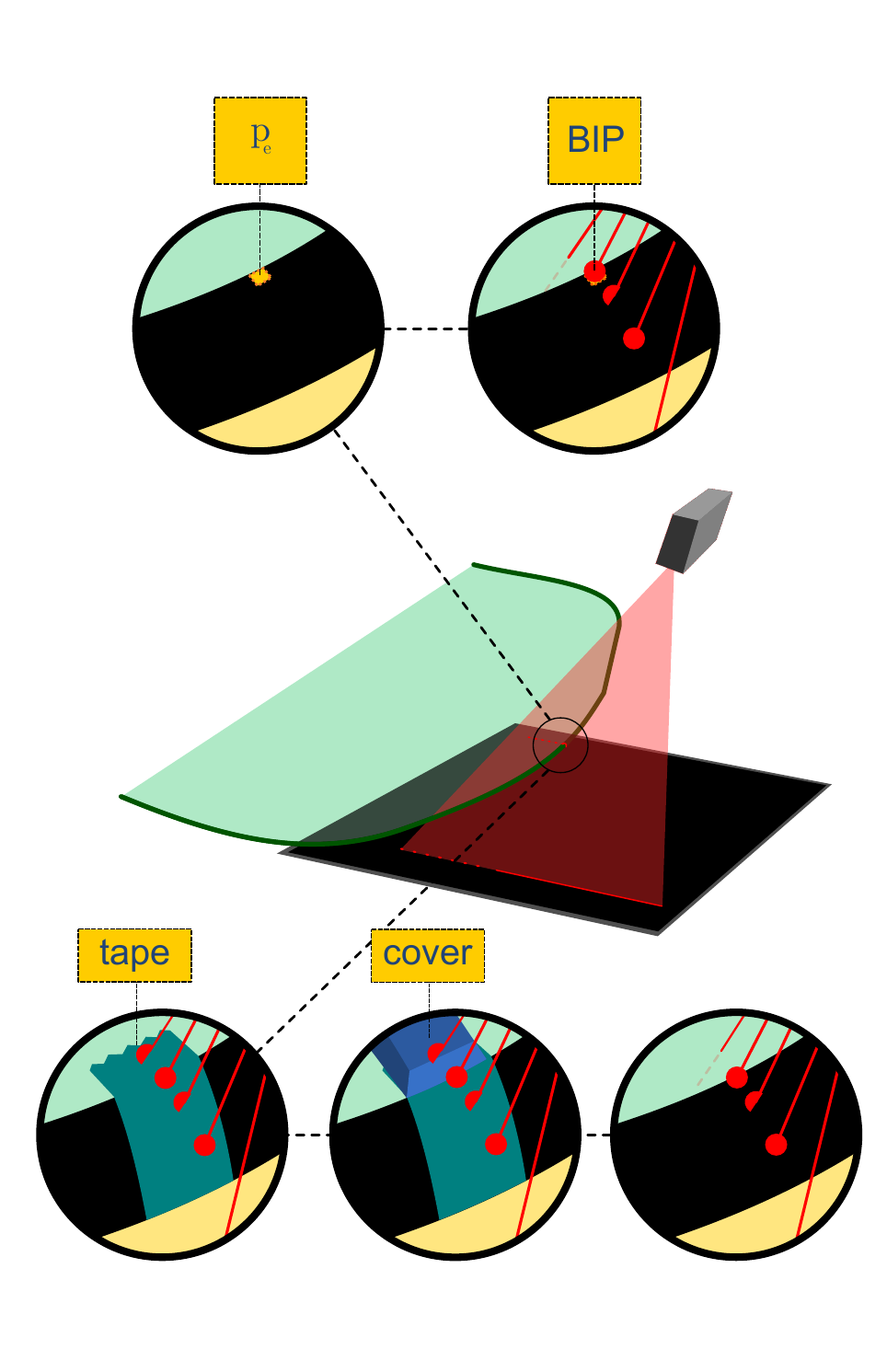}
  \caption{First row: At each scan pose, point $\boldsymbol{p}_e$ signifies the
    point location along the border height that we intend to
    capture. BIP refers to the actual laser point reading that we
    select among the laser line points. Second row: Illustration of
    the scanning operation. Third row: To assess the likelihood that
    a laser point falls on $\boldsymbol{p}_e$, we compare laser readings from three
    experiments at various angles: First, tape is laid along the
    border height. Second, a cover is placed on the surface to
    indicate the start of border boundary. Third, without any cover.}
  \label{fig:fig3-3}
\end{figure}

There is, however, the uncertainty whether laser line projection
could return points from the edge line reliably. To understand this, we
devised an experiment to assess the distribution of laser line on the
transparent-opaque surface intersection as seen in the bottom of Figure
\ref{fig:fig3-3}:
\begin{enumerate}
\item Tape is laid along the border and onto the glass surface. The
  laser reading obtained this way reflects the true contour of the
  glass.
  \item A border cover object is placed on the glass surface that
    coincides with the edge line. Another laser reading is then
    obtained.
  \item After removing the tape and cover object, another laser reading is obtained.
  \item Repeat the experiment for various scan angles.
\end{enumerate}

\begin{figure}[htp]
  \centering
  \begin{subfigure}{\columnwidth}
    \centering
    \includegraphics[width=\linewidth]{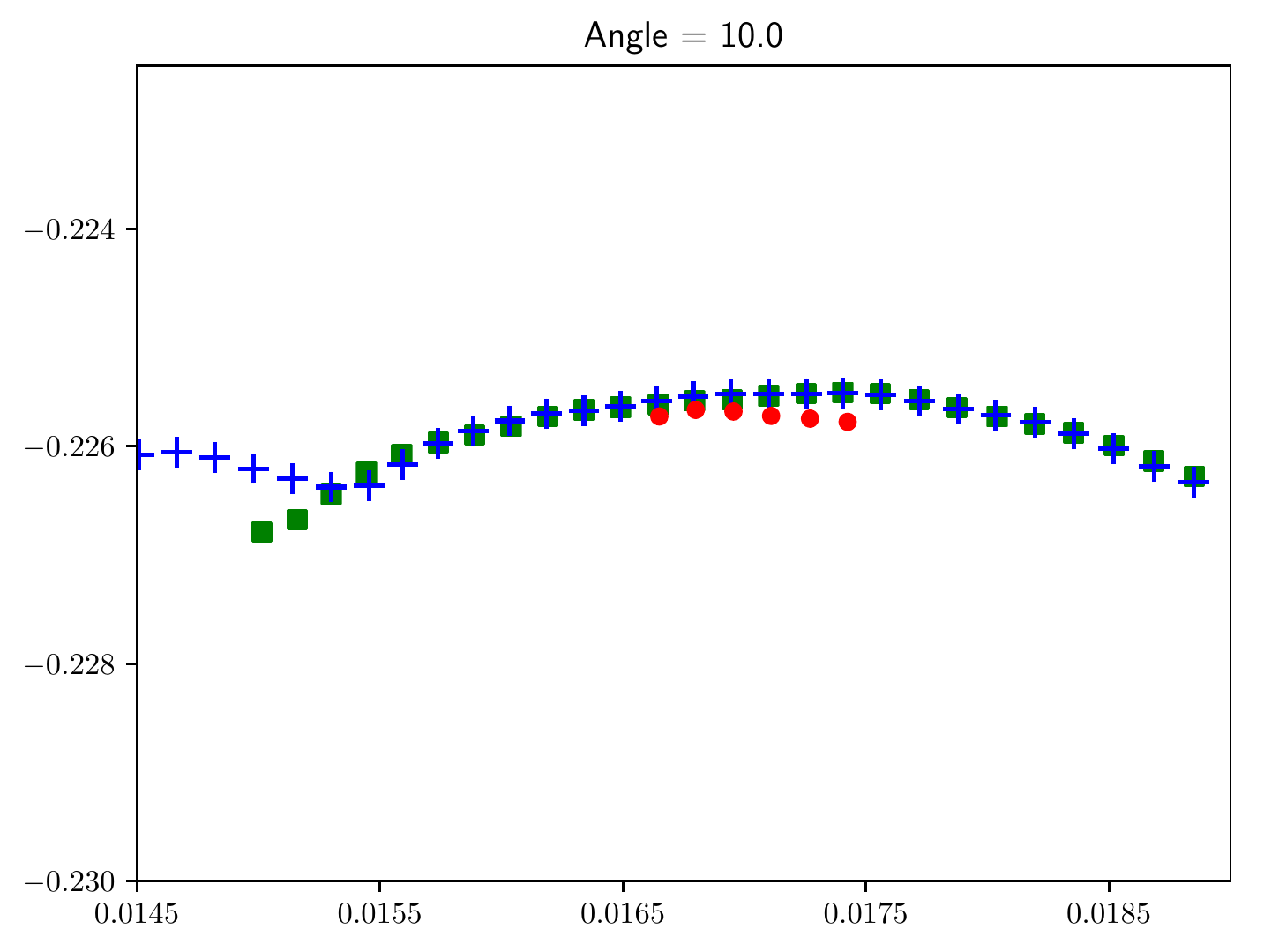}
    \caption{}
    \label{fig:fig3-4a}
  \end{subfigure}
  \begin{subfigure}{\columnwidth}
    \centering
    \includegraphics[width=\linewidth]{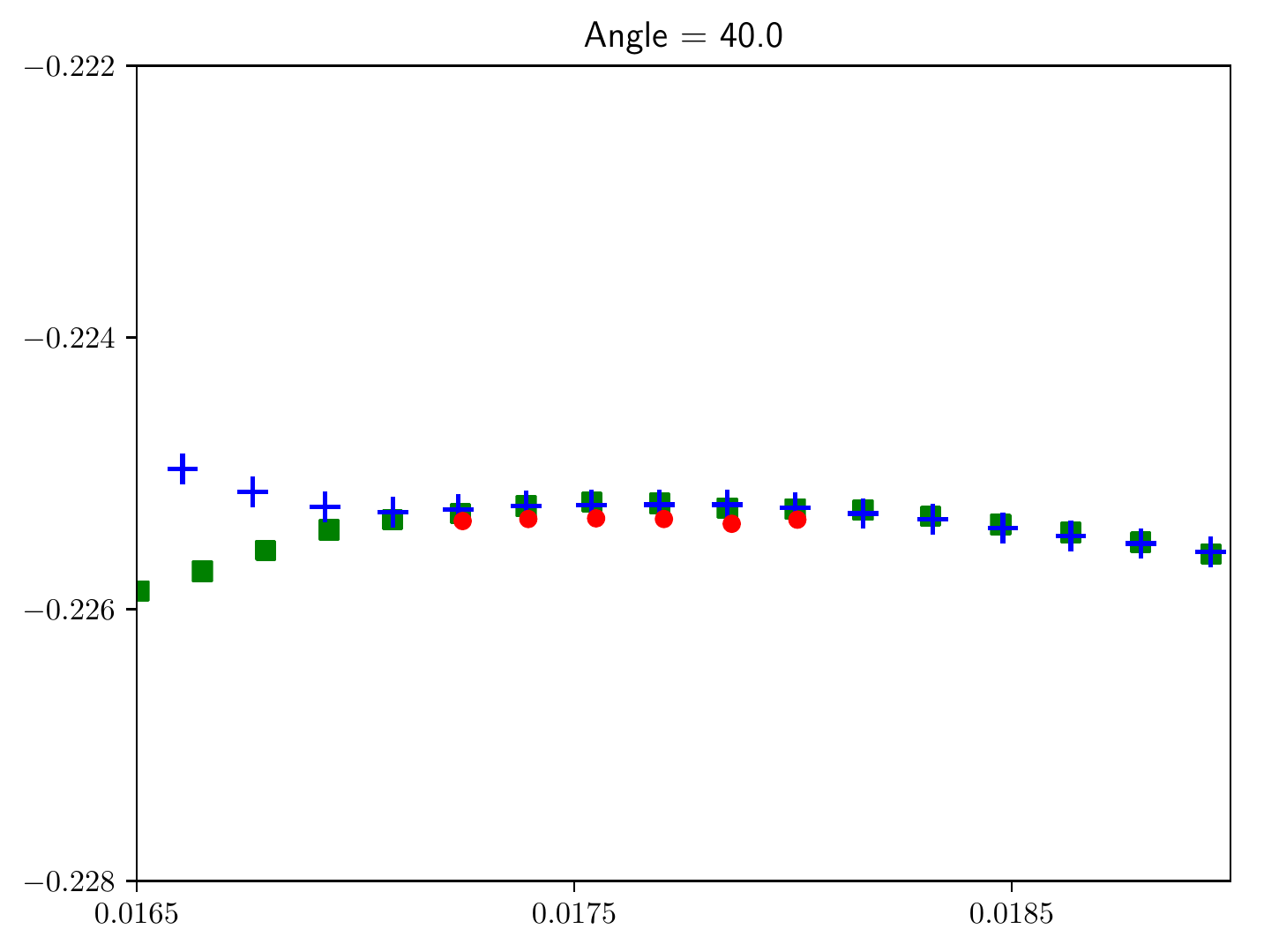}
    \caption{}
    \label{fig:fig3-4b}
  \end{subfigure}
  \caption{At different scanning angles, the recorded laser readings
    vary in terms of proximity to the desired $\boldsymbol{p}_e$. At (a) 10 degrees
    angle, the laser readings are further away from the border boundary
    as compared to scanning at (b) 40 degrees angle. Blue crosses, green
    squares and red dots refer to points captured from the surfaces of border
    cover, tape and glass edge respectively.}
  \label{fig:fig3-4}
\end{figure}

In both Figures \ref{fig:fig3-4a} and \ref{fig:fig3-4b}, the laser
reading from Step 1 and 2 are traced by the blue crosses and green
squares respectively. The first intersection from the left between the
blue and green plots corresponds to the point on the edge line and the
laser line readings obtained from Step 3 are visualised as red
plots. It was found that the selection of scanning angle affects the
degree at which the laser points coincide with the edge line. In
Figure \ref{fig:fig3-4b}, laser points from a 40 degrees scanning angle
above the horizontal plane display closer proximity to the edge line
compared to laser points from a 10 degrees scanning angle in Figure
\ref{fig:fig3-4a}.

\section{Glass pose estimation pipeline}\label{part:pipeline}

\subsection{Pipeline}

Based on the observations of the previous section, we propose the
following pipeline:
\begin{enumerate}
\item \emph{Scanning}: Sample $n$ points along the object border by
  looking at the surface with the LRF from an angle of approximately
  45 degrees. We assume in this paper that a coarse localisation has
  been performed beforehand, using existing
  methods\cite{Hata1996}\cite{Ben-Ezra2003};
\item \emph{Registration}: Perform Iterative Closest Point (ICP)
  registration between the $n$ points against an available 3D model;
\item \emph{Manipulation}: Based on the object pose estimate, compute
  the robot trajectory to perform the manipulation (here: following
  the object contour mimicking a glue dispensing task).
\end{enumerate}

\subsection{Scanning}
% Scanning method
We position the laser scanner at $n$ different poses along the glass
border with the laser line intersecting the edge. At each pose, the
scanner receives a set of points from the laser line being reflected
off the glass edge. We then proceed to filter these collinear points
to obtain a single point which we call the Border Identifier Point
(BIP). The BIP filtering method should ensure consistent point
location among the collinear points for all the $n$ poses. For our
application, we ensure that the BIP selected is the highest point
among the collinear points above the ground plane. This point
corresponds to the point $\boldsymbol{p}_e$ on the intersection between the
transparent glass surface and diffuse glass edge (refer to the
illustration on top of Figure \ref{fig:fig3-3}). Hence, in the end we
obtain set $S$ containing $n$ BIPs from $n$ scanner configurations.

\subsection{Registration}
Our proposed method requires a 3D model of the glass object to be
available. If unavailable, pre-measures can be employed to augment the
3D model construction. For instance, coating materials such as talcum
powder can be applied on the surface to obtain accurate scan
measurement. Construction of the 3D model need only be done once and
hence will not affect subsequent glue-dispensing operation time.

\begin{figure}[htp]
  \centering
  \begin{subfigure}{0.5\columnwidth}
    \centering
    \includegraphics[width=\linewidth]{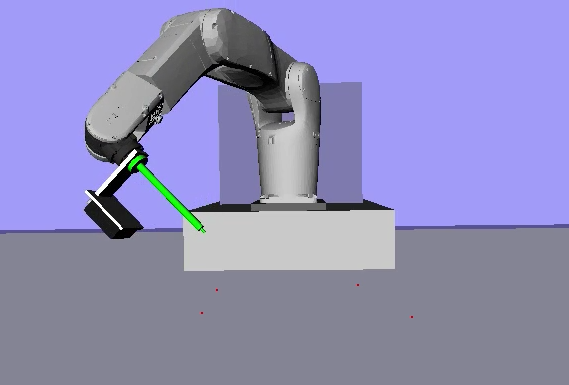}
    \caption{}
    \label{fig:fig4-0a}
  \end{subfigure}%
  \begin{subfigure}{0.5\columnwidth}
    \centering
    \includegraphics[width=\linewidth]{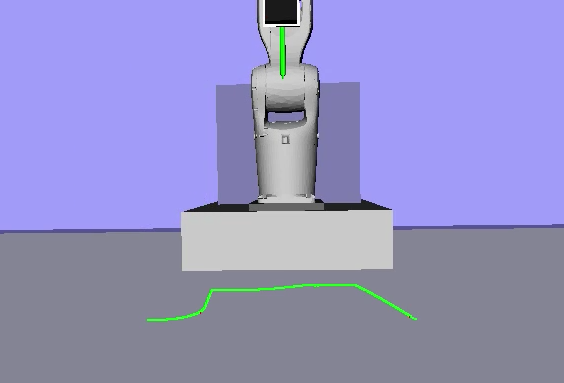}
    \caption{}
    \label{fig:fig4-0b}
  \end{subfigure}
  \caption{OpenRAVE \cite{diankov_thesis} simulation of the pose
    estimation. (a) Four scanned BIPs shown as red points. (b)
    Estimated reachable glass border segment $M'$ post-ICP in green.}
  \label{fig:fig4-0}
\end{figure}

ICP registration \cite{Holz2015} is then performed to obtain the
estimated pose of the car window. The ICP algorithm works in principle
by matching two cloud points, in this case the cloud points formed
from $S$ against the constructed 3D model, $M$. Through minimizing the
distance distribution, the algorithm returns a transformation that
aligns the two models as seen in Figure \ref{fig:fig4-0}.

\section{Experiment 1: point localisation}\label{part:experimentA}

The experiment can be briefly summarised as follows: The robot aligns
the laser line with the point-of-interest before moving the tooltip to
touch it. The resulting distance from the point-of-interest to the
final tooltip position is then recorded.

We perform this experiment on two different object profiles for
comparison: Firstly on glass edge and later on an opaque object's
edge with diffuse reflection. Through comparison of the two results,
we can obtain the relative accuracy of scanning on glass edge against
opaque object edge as ground truth.

The laser scanner used is a 2D LRF, Microepsilon Scancontrol 2600-100,
with 658 nm red semiconductor laser as light source. The LRF
transformation parameters relative to the robot's flange is calibrated
based on method described in SCALAR \cite{Lembono2018}.

\subsection{Method}

The experiment steps are as follows:
\begin{enumerate}
\item Position the laser beam line to intersect the object edge at $n$
  multiple locations spanning the perimeter and calculate the BIP
  point $^{B} \boldsymbol{p}_\mathrm{BIP}$ at each location, where $B$
  represents the robot's base coordinate frame.
\item At each location, mark the point on the object under the laser
  beam line which is supposed to coincide with the respective
  $^{B} \boldsymbol{p}_\mathrm{BIP}$ and move the tooltip to touch it while
  keeping a predefined orientation of the tooltip,
  $\boldsymbol{R}_\mathrm{approach}$. Record the joint states when the tooltip touches the
  desired point as $\boldsymbol{q}_\mathrm{ref}$.
\item Program the robot to move towards each
  $^{B} \boldsymbol{p}_\mathrm{BIP}$ with $\boldsymbol{R}_\mathrm{approach}$ as approach
  orientation. Record the resulting joints states as $\boldsymbol{q}_\mathrm{actual}$.
\end{enumerate}

Let the Forward Kinematics mapping from the
\textit{configuration-space} to \textit{task-space} be $FK :
\boldsymbol{q} \mapsto \boldsymbol{p}$ and
the tooltip transformation relative to the robot's flange be
$^{\mathrm{flange}} \boldsymbol{T}_{\mathrm{tooltip}}$.

The localisation error can be defined as follows
\begin{equation}
  \epsilon := || FK(\boldsymbol{q}_\mathrm{actual})\ ^\mathrm{flange}\boldsymbol{T}_\mathrm{tooltip} - FK(\boldsymbol{q}_\mathrm{ref})\ ^\mathrm{flange}\boldsymbol{T}_\mathrm{tooltip}. ||\nonumber
\end{equation}

One can then define the mean and maximum errors across all tested points.

\subsection{Results and discussion}

The table below compares the performance result for the two object profiles:

\begin{table}[H]
  \caption{Point localisation error}
  \setlength{\tabcolsep}{7pt}
  \centering
  \begin{tabular}{lcc}\toprule
    & Opaque object & Glass \\\midrule
    Mean error ($\pm$ std) (mm) & 1.01 ($\pm$ 0.14) & 0.94 ($\pm$ 0.28)\\
    Max error (mm) & 1.27 & 1.35 \\
    \bottomrule
  \end{tabular}
  \label{tab:3}
\end{table}

The two experimental results show minimal discrepancy suggesting
similar performance in point localisation. Such an outcome provides
validation that scanning on glass edge offers comparable reliability
to scanning on opaque objects. Henceforth, we can make use of glass
edge measurement reading as input to glass pose estimation.

\section{Experiment 2: full pipeline}\label{part:experimentB}

\begin{figure}[htp]
  \centering
  \includegraphics[width=\columnwidth]{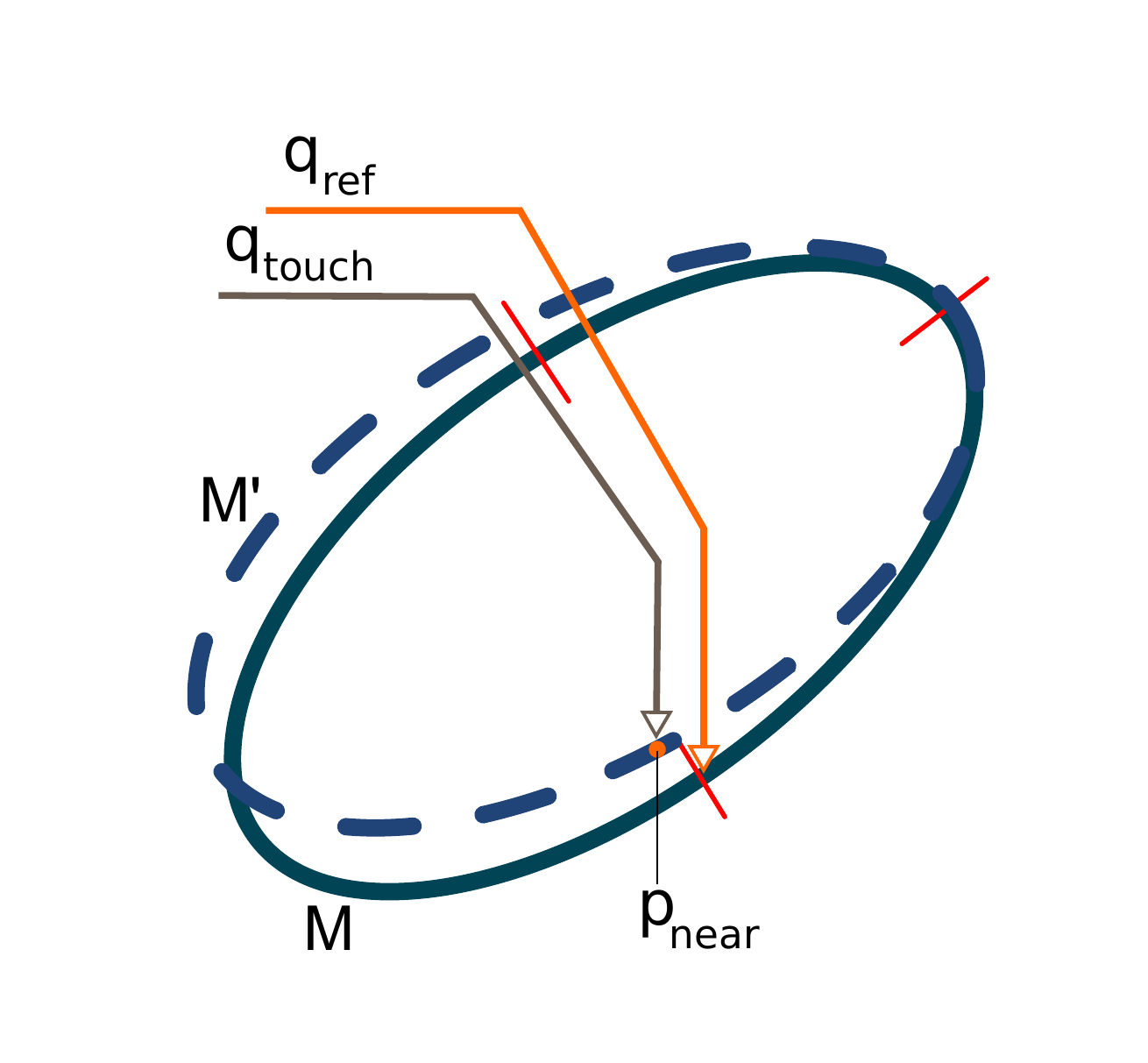}
  \caption{For every scan pose, we move the robot's tooltip manually
    to touch the $\mathrm{BIP}$ point on the real object $M$ and record the
    joint states $\boldsymbol{q}_\mathrm{ref}$.
    After ICP, we find the nearest point $\boldsymbol{p}_\mathrm{near}$ on
    the estimated $M'$ to the tooltip point at $\boldsymbol{q}_\mathrm{ref}$ and move
    towards it. The resulting joint states $\boldsymbol{q}_\mathrm{touch}$ is
    recorded. The tooltip positions at $\boldsymbol{q}_\mathrm{ref}$ and $\boldsymbol{q}_\mathrm{touch}$ are
    then measured to assess the task accuracy.}
  \label{fig:fig5}
\end{figure}

To validate the accuracy of the full task pipeline, an experiment is
devised to measure the accuracy of the transparent object pose
estimation. For visual guidance, we refer the reader to Figure
\ref{fig:fig5}.

\subsection{Method}

The following validation steps are performed following the steps in
Section \ref{part:experimentA}:
\begin{enumerate}
\item Perform ICP on the previously obtained $n$
  $^{B} \boldsymbol{p}_\mathrm{BIP}$ points to obtain $^M \boldsymbol{T}_{M'}$
  transformation matrix that transform $M$ to the estimated pose $M'$
  in the workspace.
\item For every $FK(\boldsymbol{q}_\mathrm{ref})^\mathrm{flange}\boldsymbol{T}_\mathrm{tooltip}$, we calculate the
  nearest point in $M'$ as $^{B} \boldsymbol{p}_\mathrm{near}$.
\item Program the robot to move towards $^{B} \boldsymbol{p}_\mathrm{near}$
    with $\boldsymbol{R}_\mathrm{approach}$ as approach orientation. Record the resulting
    joint states as $\boldsymbol{q}_\mathrm{touch}$.
\end{enumerate}

The localisation error is defined as
\begin{equation}
  \epsilon := || FK(\boldsymbol{q}_\mathrm{touch})\ ^\mathrm{flange}\boldsymbol{T}_\mathrm{tooltip} - FK(\boldsymbol{q}_\mathrm{ref})\ ^\mathrm{flange}\boldsymbol{T}_\mathrm{tooltip}. ||\nonumber
\end{equation}

One can then define the mean and maximum errors across all discretised
points along the glass border.

\subsection{Results and discussion}

In the following Table \ref{tab:2}, we list down the average operation
time for each stage making up the full glue-dispensing operation with $n$=4:

\begin{table}[H]
  \caption{Average operation time}
  \setlength{\tabcolsep}{14pt}
  \centering
  \begin{tabular}{lc}\toprule
    Stage & Average time (s)\\\midrule
    Scanning & 14.845\\
    Pose estimation & 0.023\\
    Path planning & 0.782\\
    Execution & 26.884\\
    \bottomrule
  \end{tabular}
  \label{tab:2}
\end{table}

While scanning and execution stages contribute significantly to the
total operation time, the values presented in the table do not reflect
the best time achievable as they are highly dependent on the robot's
velocity. In our experiment, we limit the velocity and acceleration of
the robot for safety reason.

The table below presents the pose estimation result based on the
validation metric mentioned previously with $n$=12:

\begin{table}[H]
  \caption{Full pipeline localisation error}
  \setlength{\tabcolsep}{10pt}
  \centering
  \begin{tabular}{lc}\toprule
    & Localisation error \\\midrule
    Mean error ($\pm$ std) (mm) & 0.88 ($\pm$ 0.39)\\
    Max error (mm) & 1.65 \\
    \bottomrule
  \end{tabular}
  \label{tab:4}
\end{table}

\section{Conclusion}\label{part:conclusion}

In this paper we have looked into the feasibility of accurately
locating transparent object using a 2D LRF with the ultimate goal of
having a 6 DOF robot manipulating on it. Based on a physical
understanding of laser-glass interactions, we introduced a method
that combines 6 DOF robotic arm with 2D LRF to estimate a transparent
object pose in the workspace by looking at the object edge. The
method was tested in an industrial scenario which emphasises
the need for fast and accurate online operation. Experiments showed
that our method could achieve errors in the order of the
millimetre, making it suitable for many industrial scenarios, such as
manipulation or glue dispensing. 

Future work will include integration with upstream coarse
localisation. The accuracy can also be further improved by performing
a more complete calibration of laser-in-hand robotic system\cite{Lembono2018}.

\section*{Acknowledgment}
This work was supported in part by NTUitive Gap Fund NGF-2016-01-028 and SMART Innovation Grant NG000074-ENG.

%\addtolength{\textheight}{-12cm}

\bibliography{gluebot}
\bibliographystyle{ieeetr}
\end{document}